\documentclass[11pt]{article} 
\usepackage{rldmsubmit,palatino}
\usepackage{graphicx}
\usepackage[colorlinks=true,linkcolor=blue,citecolor=blue,urlcolor=blue]{hyperref}
\usepackage{amscd,amsfonts,amssymb,amstext,latexsym} 
\usepackage{amsmath,mathbbol,mathrsfs,stmaryrd, mathtools} 
\usepackage[numbers,sort&compress]{natbib}
\usepackage{caption}
\usepackage{subcaption}
\usepackage{algorithm}
\usepackage{algpseudocode}
\usepackage{todonotes}
\usepackage{microtype}
\usepackage{amsthm}
\usepackage{wrapfig}
\usepackage[colorlinks=true,linkcolor=blue,citecolor=blue,urlcolor=blue]{hyperref}
\newtheorem{theorem}{Theorem}
\newcommand{\awr}{
\begin{algorithmic}
\Require Basis set $\mathbf{F}$, $s$, $\delta_t$, $\rho(\phi)$, $T (\phi)$, splitting tolerance $\tau_s$.

\ForAll {Functions $\phi$ activated by state $(s, a)$}
\State Update $\rho(\phi)$, $O(\phi)$, $T(\phi)$ \&  $\rho(\phi_c), T(\phi_c)$ $\forall$ children $\phi_c$ of $\phi$.
\EndFor
\If {$C(\phi) \geq \tau_s$ and $C(\phi) = \max_k  (C(\phi_k)) $}

    \State Replace $\phi$ with its children in dimension $d$, $\phi^d_{j, k}$, where $d$ maximises the average relevance of the children of $\phi$.
\EndIf
\end{algorithmic}
}

\newcommand{\ibfdd}{
\begin{algorithmic}
\Require  Basis set $\mathbf{F}$, $s$, $\delta_t$, $\rho(\phi)$, $T (\phi)$, combination tolerance $\tau_c$.
\ForAll {$\phi_f = \phi_g \phi_h$ s.t. $\phi_g, \phi_h \in \mathbf{F}, \phi_f \notin \mathbf{F}$ and $\phi_f(s) \neq 0$}
\State Update $\rho(\phi_f), T(\phi_f)$
\EndFor
\If {$| \rho(\phi_f) | \geq \tau_c$ and $| \rho(\phi_f) | = \max_k | \rho(\phi_k) |$}
    \State $\mathbf{F} \leftarrow \mathbf{F} \cup \{ \phi_f \}$
\EndIf
\end{algorithmic}}
\title{Adaptive Online Value Function Approximation with Wavelets}
\author{
Michael Beukman\\
School of Computer Science and Applied Mathematics\\
University of the Witwatersrand\\
Johannesburg, South Africa \\
\texttt{michael.beukman1@students.wits.ac.za} \\
\And
Michael Mitchley\\
School of Computer Science and Applied Mathematics\\
University of the Witwatersrand\\
Johannesburg, South Africa \\
\texttt{barcoded@gmail.com}
\And
Dean Wookey\\
School of Computer Science and Applied Mathematics\\
University of the Witwatersrand\\
Johannesburg, South Africa \\
\texttt{wookey.dean@gmail.com}
\And
Steven James\\
School of Computer Science and Applied Mathematics\\
University of the Witwatersrand\\
Johannesburg, South Africa \\
\texttt{steven.james@wits.ac.za}
\And
George Konidaris\\
Department of Computer Science\\
Brown University\\
Providence RI, 02912 \\
\texttt{gdk@cs.brown.edu}
}

\newcommand{\basis}{\mathbf{\Phi}}
\newcommand{\weights}{\mathbf{w}}

\newtheoremstyle{mytheoremstyle}{0pt}{0pt}{\itshape}{}{\bfseries}{.}{.5em}{} 

\theoremstyle{mytheoremstyle}

\usepackage{xpatch}
\makeatletter
\xpatchcmd{\proof}{\topsep6\p@\@plus6\p@\relax}{}{}{}
\makeatother

\usepackage{thmtools}

\declaretheoremstyle[%
  spaceabove=-6pt,%
  spacebelow=0pt,%
  headfont=\normalfont\itshape,%
  postheadspace=1em,%
  qed=\qedsymbol%
]{mystyle} 
\declaretheorem[name={Proof},style=mystyle,unnumbered,
]{prf}

\begin{document}

\maketitle

\begin{abstract}

Using function approximation to represent a value function is necessary for continuous and high-dimensional state spaces.
Linear function approximation has desirable theoretical guarantees and often requires less compute and samples than neural networks, but most approaches suffer from an exponential growth in the number of functions as the dimensionality of the state space increases.
In this work, we introduce the wavelet basis for reinforcement learning.
Wavelets can effectively be used as a fixed basis and additionally provide the ability to adaptively refine the basis set as learning progresses, making it feasible to start with a minimal basis set. This adaptive method can either increase the granularity of the approximation at a point in state space, or add in interactions between different dimensions as necessary. 
We prove that wavelets are both necessary and sufficient if we wish to construct a function approximator that can be adaptively refined without loss of precision. 
We further demonstrate that a fixed wavelet basis set performs comparably against the high-performing Fourier basis on Mountain Car and Acrobot, and that the adaptive methods provide a convenient approach to addressing an oversized initial basis set, while demonstrating performance comparable to, or greater than, the fixed wavelet basis.
To aid in reproducibility, we publicly release our source code.\footnote{\url{https://github.com/Michael-Beukman/WaveletRL}}

\end{abstract}


\acknowledgements{Computations were performed using High Performance Computing infrastructure provided by the Mathematical Sciences Support unit at the University of the Witwatersrand.}

\startmain 

\section{Introduction}

Representing a value function in reinforcement learning (RL) in continuous state spaces requires function approximation. 
One approach is non-linear function approximation, where a neural network is trained to learn useful features from raw observations. 
However, this class of methods requires many samples, large amounts of computation~\citep{drl_survey}, and does not possess theoretical or convergence guarantees~\citep{sutton2018reinforcement}.
Another common scheme is \textit{linear function approximation}, where the value function is approximated by a weighted sum of non-learnable basis functions.
This results in simple algorithms and an error surface convex in the weights, but still allows for the representation of complex value functions because the basis functions themselves can be arbitrarily complex.
The obvious question is then: which basis functions should one use?

Several fixed basis schemes have been introduced (e.g., \citep{powell1987radial,fourier_basis,albus1971tiles}), all with the inherent disadvantage of a combinatorial explosion that makes them unsuited to high-dimensional domains. 
Consequently, research has focused on either constructing basis functions from data in both discrete and continuous state spaces, or selecting an appropriate set from a fixed dictionary of candidate basis functions with the aim of producing basis function sets that are subexponential in the number of dimensions. 
A third approach is \textit{adaptive methods} (e.g., \citep{whiteson2007adaptive,lin2010evolutionary,geramifard2011online,li2019Adaptive}), which are a hybrid of the above. 
Adaptive methods create new basis functions that complement or replace an existing set of basis functions on which learning is performed. 
They have the advantage in that they are online, which reduces the sample and computational complexity, and do not require a large dictionary of candidate functions, since they add representation complexity incrementally.

We extend existing adaptive methods to basis functions based on \textit{wavelets}, which are able to approximate functions at various scales and locations, and which can be refined to build a more accurate representation where such detail is necessary. 
We further prove that wavelets are both necessary and sufficient for function splitting approaches. 
We present an algorithm for value function approximation that begins with a minimal set of basis functions, and only adds interactions between different dimensions as required. 
We test our approach on Mountain Car and Acrobot and our results show that a fixed wavelet basis is competitive with the high-performing Fourier basis~\citep{fourier_basis}. 
Furthermore, the adaptive techniques perform comparably to the fixed basis, while not starting out with a complete basis set.

\section{Value Function Approximation}

The reinforcement learning problem in continuous domains is typically modelled as a Markov Decision Process and described by a tuple $(S, A, P, R, \gamma)$, where $S \subseteq \mathbb{R}^d$ is a $d$-dimensional state space, $A$ is a set of actions, $P(s' |s, a)$ describes the probability of transitioning to state $s'$ after having performed action $a$ in state $s$, $R(s, a)$ describes the reward received for executing such a transition, and $\gamma$ is the discount factor. The agent is required to learn a policy $\pi$ mapping states to actions that maximises the return (discounted future sum of rewards) at time $t$: $G_t = \sum_{i=0}^{\infty} \gamma^i R(s_{t+i}, a_{t+i})$ ~\citep{sutton2018reinforcement}.
Given a policy $\pi$, RL algorithms often estimate a \textit{value function}: $V^\pi(s) = \mathbb{E}_\pi \big[ \sum_{i=0}^{\infty} \gamma^i R(s_{i}, a_{i}) | s_0 = s \big]$, where action $a$ is selected according to policy $\pi$. Linear value function approximation represents $V^\pi$ as weighted sum of $n$ basis function $\basis$: $V^\pi(s)  \approx \weights \cdot \basis = \sum_{i=1}^{n} w_i \phi_i(s)$.
This approximation is linear in the components of the parameter (or weight) vector, $\weights$, which results in simple update rules and a quadratic error surface.
\paragraph*{Basis Functions}
In function approximation, a \textit{basis} is a set of orthonormal functions spanning a function space, such that anything within that function space can be exactly reconstructed using a unique weighted sum of the basis functions. 
Typically, we are interested in orthonormal bases for $\mathcal{L}_2(\mathbb{R})$, the space of all finite square-integrable functions. In Euclidean spaces, $f(x) \in \mathcal{L}_2(\mathbb{R})$ implies $\int_{-\infty}^\infty f^2(x) dx$ is finite, a property satisfied by all value functions with finite return. Formally, if a basis is overcomplete (i.e. not all functions are orthogonal), then it is known as a \textit{frame}.

A simple choice of basis function is \textit{tile coding}~\citep{albus1971tiles}, where the state space is discretised into tile basis functions that evaluate to $1$ within a fixed region, and $0$ elsewhere.
Although tile coding forms a basis, it restricts the value function to be piecewise constant. 
Another approach is the \textit{polynomial basis} \citep{lagoudakis2003least}, which sets $\phi_i(s) = \Pi_{j=1}^d s_j^{c_{i,j}}$, where each $c_{i,j} \in [0, \ldots, n]$ and $n$ denotes the order of the basis. While orthonormal polynomials exist, simple ones are not, and may lead to redundancies in their representations. 
A more common scheme is \textit{radial basis functions} (RBFs)~\citep{powell1987radial}, where each function is a Gaussian with a given mean and variance. 
RBFs have local support and are therefore well-suited to representing value functions with local discontinuities. 
The \textit{Fourier basis}~\citep{fourier_basis} uses Fourier series terms as basis functions, setting $\phi_i(s) = \cos (\pi \mathbf{c}_i \cdot s)$ where for order $n$, $\mathbf{c}_i = [c_1, \ldots, c_d]$ is a vector of coefficients between $0$ and $n$.
The Fourier basis requires choosing just a single parameter (the order) and in practice outperforms RBFs and polynomials on some common low-dimensional benchmarks~\citep{fourier_basis}.

The main shortcoming of these fixed-basis approaches is that the number of basis functions grows exponentially with the dimension of the state space.
This has led to feature selection approaches that use a fixed basis as a feature dictionary, but only select a small subset of terms to compactly represent the value function.
Ideally, such methods would allow the function approximator to represent extra detail where necessary and handle local discontinuities. 
However, the above basis function schemes are poorly suited to this: tile coding requires careful discretisation and can only represent piecewise-constant functions, RBFs lack an obvious way of setting their centres and variances, and the Fourier and polynomial bases produce functions with global support. 
We therefore require a new basis scheme suitable for general function approximation that allows us to add spatially local basis functions to incrementally add detail to the value function representation.

\section{Wavelets}
Through the use of a \textit{Fourier transform}, we can represent any periodic function as an integral of sines and cosines with varying frequencies. 
However, since sines and cosines have global support, the Fourier transform becomes cumbersome when representing \textit{transient} or local phenomena~\citep{mallat1999wavelet}.
To deal with this issue, we can instead consider \textit{wavelets}. Wavelets are simply functions $\phi: \mathbb{R} \to \mathbb{R}$, some families of which are compactly supported~\citep{daubechies_wavelets} (nonzero in a finite interval) allowing us to model local phenomena. 
There are many types of wavelets but we focus on B-Spline father wavelets here, the first three orders of which are given by:
\begin{align}
    \phi^0(x) &= 
    \left\{ \begin{array}{ll}
            1 & 0 \leq x \leq 1 \\
            0 &\text{otherwise}
    \end{array} \right. 
    &
    \phi^1(x) &= 
    \left\{ \begin{array}{ll}
            x & 0 \leq x \leq 1 \\
            2 - x & 1 \leq x \leq 2 \\
            0 &\text{otherwise}
    \end{array} \right.
    &
    \phi^2(x) &= 
    \left\{ \begin{array}{ll}
            0.5x^2 & 0 \leq x \leq 1 \\
            0.75 - (x - 1.5)^2 & 1 \leq x \leq 2 \\
            0.5(x - 3)^3 & 2 \leq x \leq 3 \\
            0 &\text{otherwise}
    \end{array} \right.
\end{align}
In practice, these functions are normalised to satisfy $\int_{-\infty}^{\infty} (\phi(x))^2dx = 1$.
We note that the $\text{zero}^\text{th}$ order B-Spline function is also referred to as a Haar wavelet \citep{haar1909theorie}, and can produce functions equivalent to disjoint tile coding.
\begin{figure}
\begin{subfigure}{0.24\linewidth}
    \centering
    \includegraphics[width=1\linewidth]{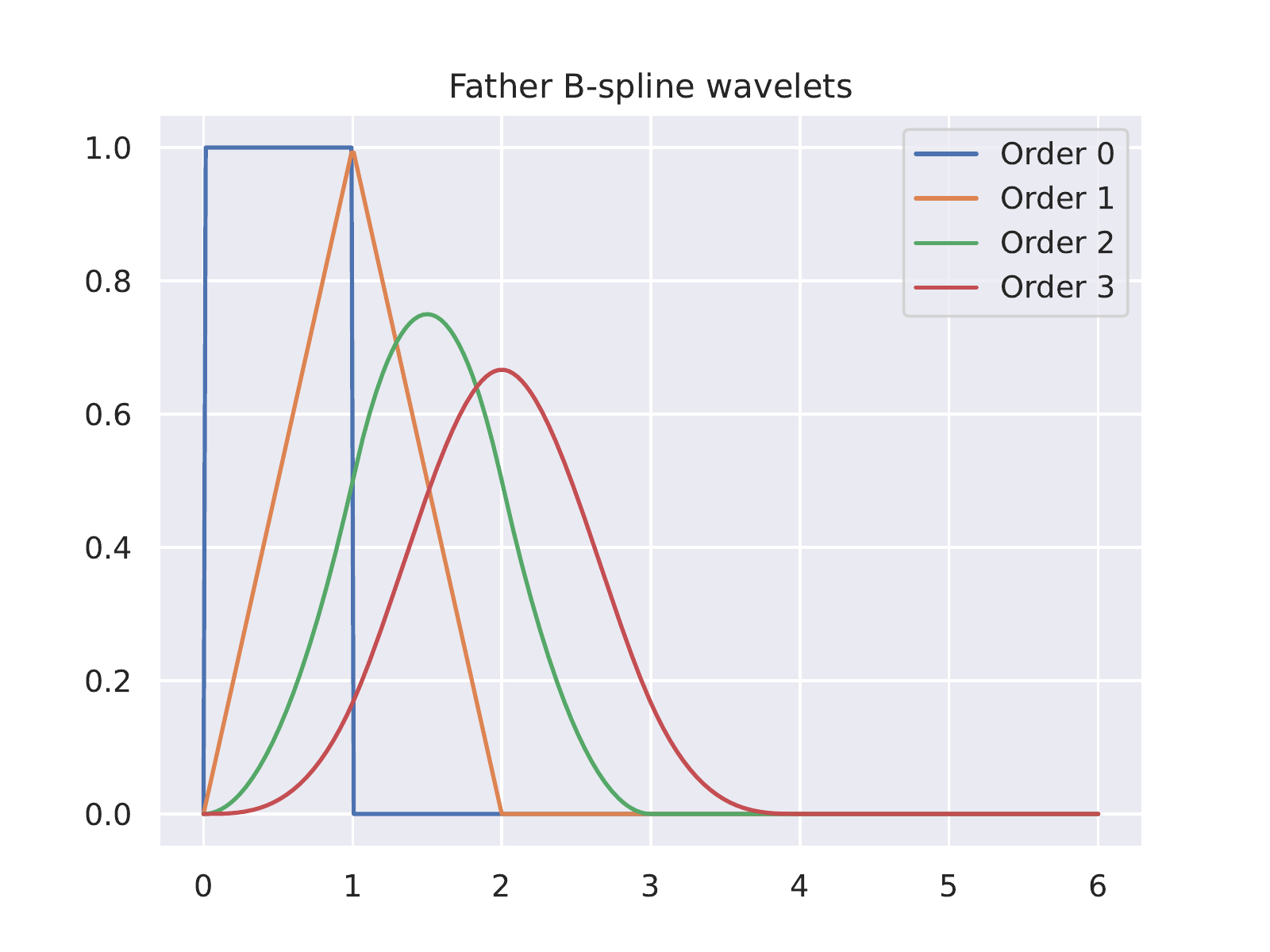}
    \caption{Different orders}
    \label{fig:bspline_order_1_to_4}
\end{subfigure}
\begin{subfigure}{0.24\linewidth}
    \centering
    \includegraphics[width=1\linewidth]{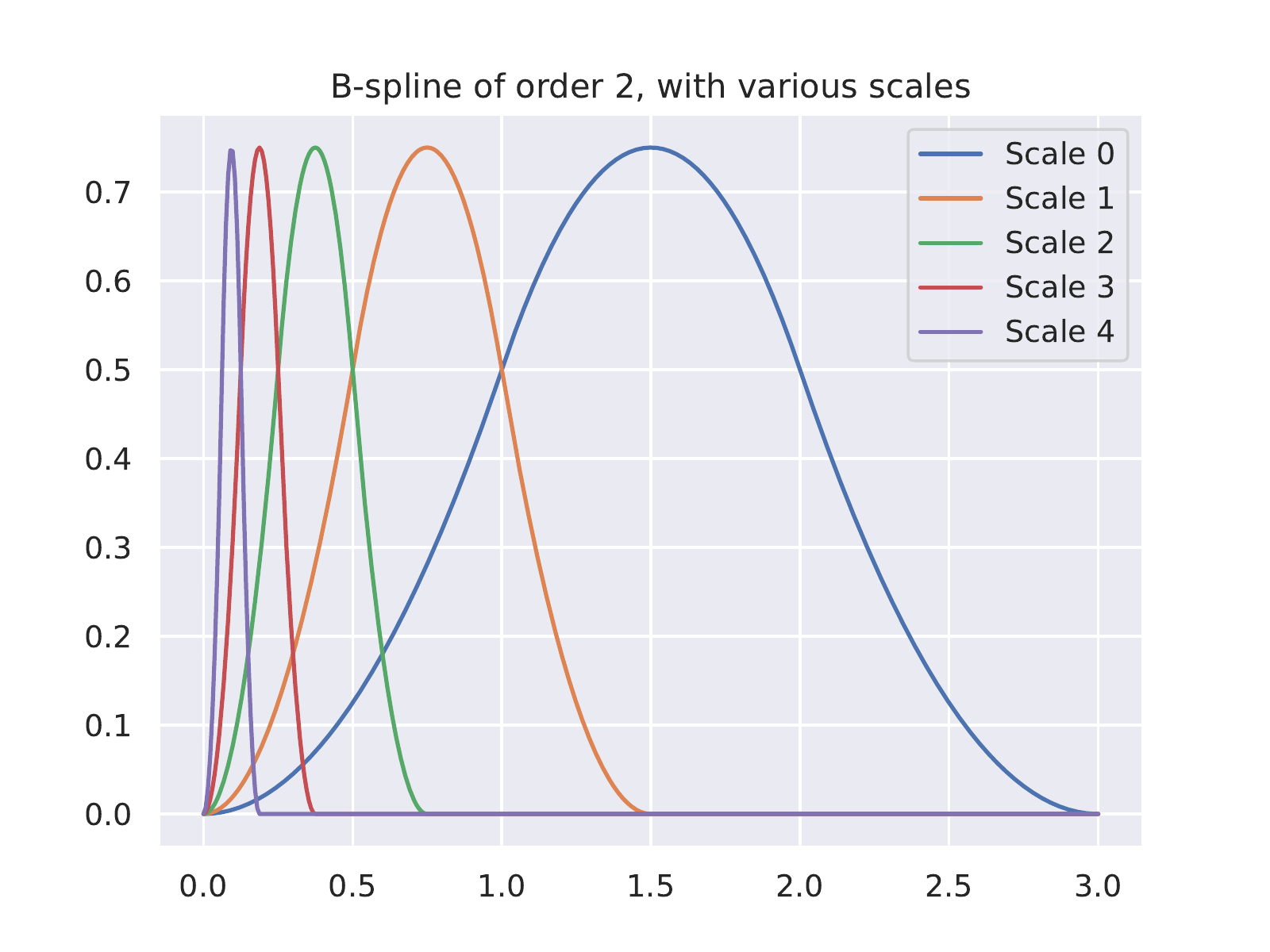}
    \caption{Scaling }
    \label{fig:bspline_scale}
\end{subfigure}
\begin{subfigure}{0.24\linewidth}
    \centering
    \includegraphics[width=1\linewidth]{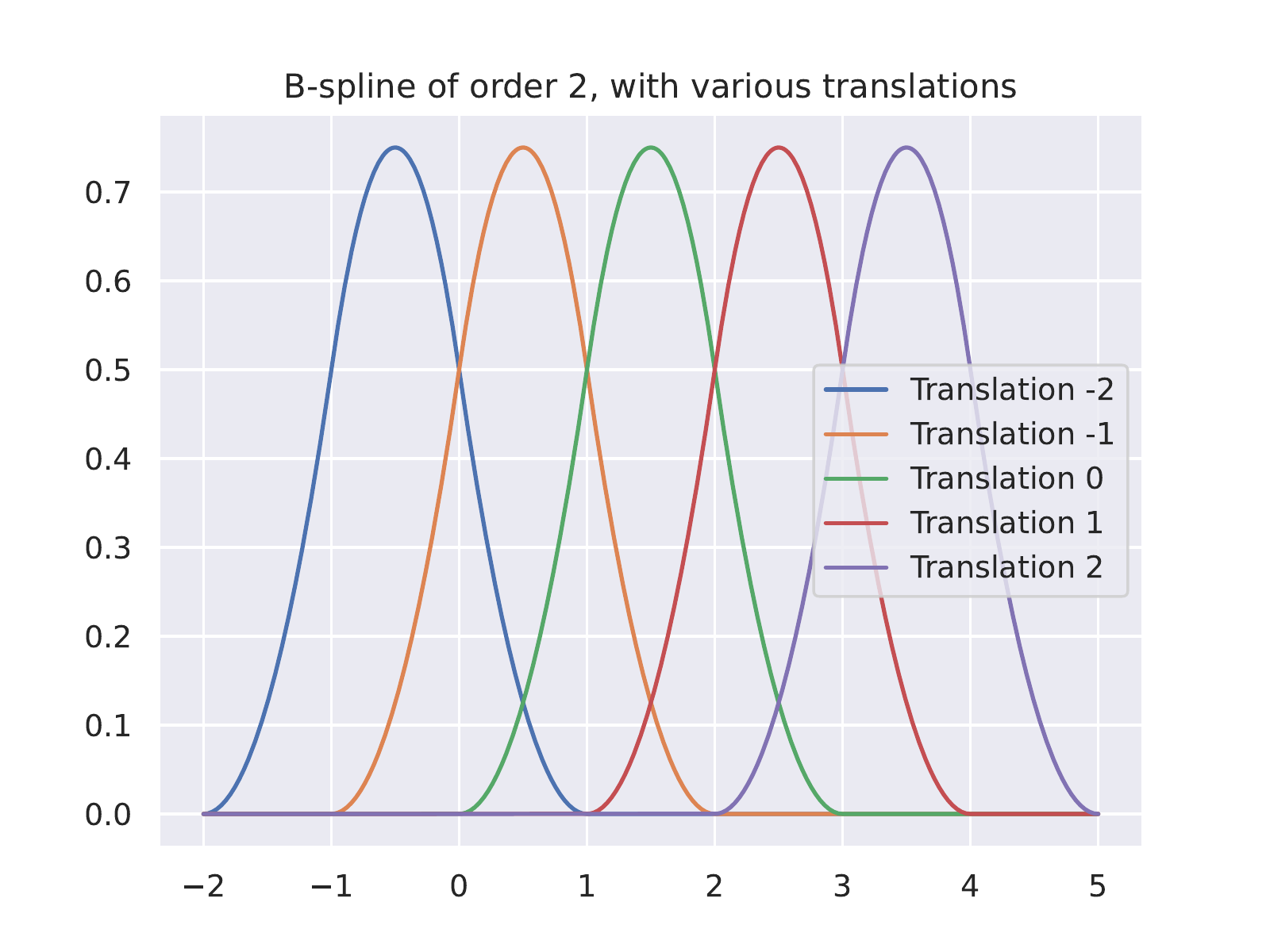}
    \caption{Translating}
    \label{fig:bspline_translation}
\end{subfigure}
\begin{subfigure}{0.24\linewidth}
    \centering
    \includegraphics[width=1\linewidth]{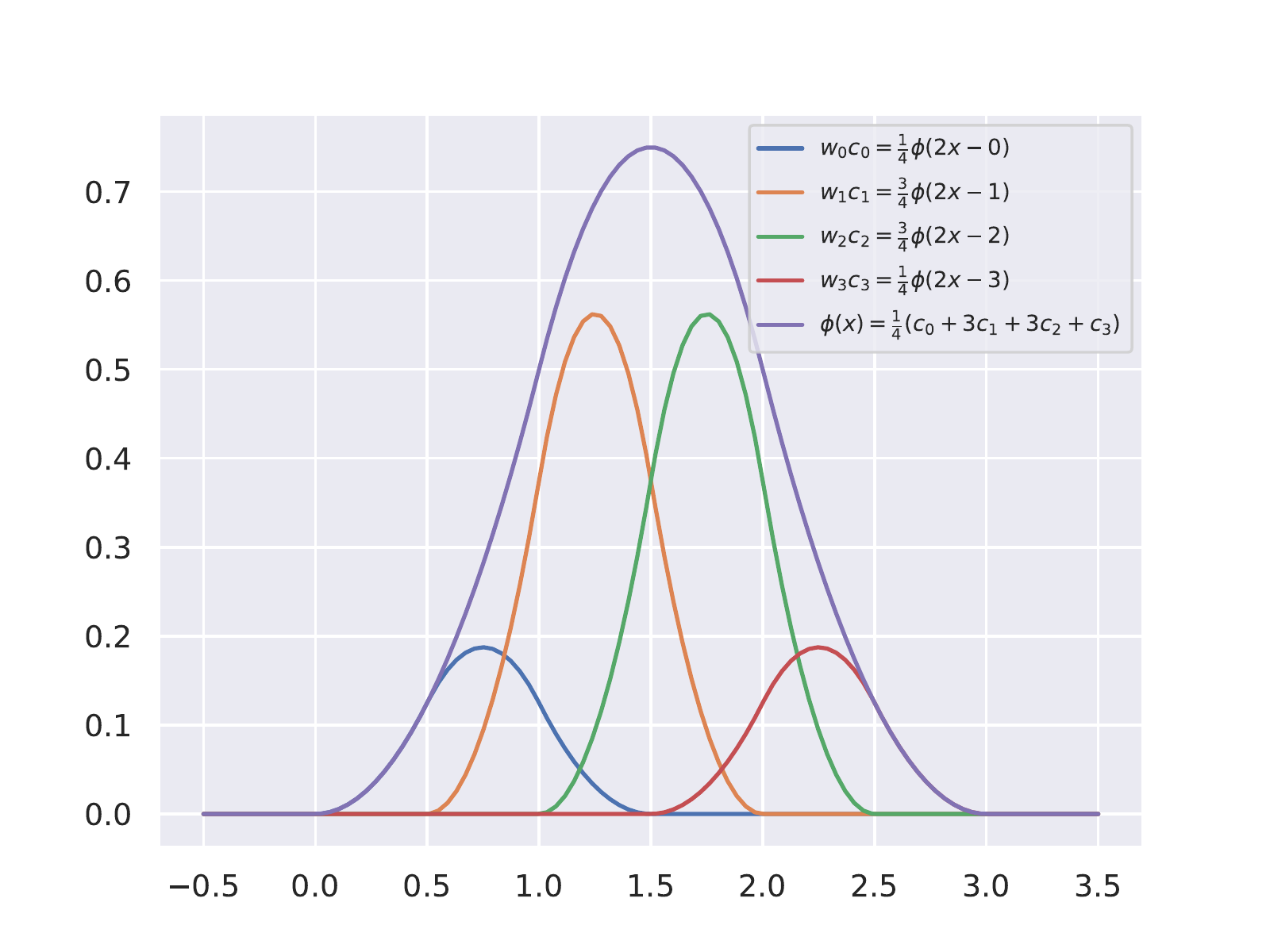}
    \caption{
    Refineability
    }
    \label{fig:spline_children_buildup}
\end{subfigure}
\caption{B-Spline wavelets of different (a) orders, (b) scales and (c) translations. In (d), the original function (purple) can be represented as a weighted sum of its ``children'' functions. For clarity, these functions are not normalised.}
\label{fig:bspline_all}
\end{figure}
We can scale (dilate) an arbitrary wavelet of order $n$ by multiplying the input $x \in \mathbb{R}$ by $2^j$ to obtain $\phi_j^n(x) = \phi^n(2^j x)$. We can further translate it by $k \in \mathbb{Z}$ to obtain $\phi_{j,k}^n(x) = \phi^n(2^j x - k)$. Some examples are shown in \autoref{fig:bspline_all}. One appealing property of wavelets is that they are \textit{refinable}, meaning that they can be replaced with a weighed sum of smaller copies of themselves at regular intervals. Concretely, as shown in \autoref{fig:spline_children_buildup}, they satisfy the \textit{refinability equation}:
\begin{align}
    \label{eq:refinability}
    \phi(x) = \sum_k w_k \phi(mx - k) & \text{ with } m > 1, k \in \mathbb{Z}
\end{align} 
We now prove that wavelets are both necessary and sufficient as a basis that obeys this equation.

\begin{theorem}
If one wishes to split a basis function $\phi(x) \in \mathcal{L}_2(\mathbb{R})$ (that is, the basis function is finite square-integrable) into a finite number of smaller copies of itself such that the value function remains unchanged, using \autoref{eq:refinability} where $m > 1$ and $k \in \mathbb{Z}$, with a finite number of $w_k$ nonzero, it is necessary and sufficient to use a compactly
supported father wavelet as $\phi$.
\end{theorem}
\begin{prf}
The above conditions define a father wavelet function~\citep{strang1989wavelets}.
Further, \autoref{eq:refinability} is obeyed by all father wavelet functions with refinement mask $m$~\citep{daubechies_wavelets}. If a father wavelet has compact support, it will have a finite dilation series.
\end{prf}

\subsection{Wavelets as a basis for linear function approximation in reinforcement learning}

To construct the wavelet basis for RL, given a state space $S \subseteq [0, 1]^d$, we choose a specific order $n$ and scale $j$. For each dimension of the state space, we create wavelets with scale $j$, order $n$ and translations $-n \leq k < 2^j$. The full basis then consists of all combinations (products) of $d$ of these functions, such that each term has a unique dimension. This is a fixed basis, providing a drop-in replacement for the methods described above. As an example, with $d=2, n = 1, j = 0$, we have state $[s_1, s_2]$ and the atomic functions are $\{\phi(s_1 - 0), \phi(s_1 + 1), \phi(s_2 - 0), \phi(s_2 + 1)\}$. All pairwise products with distinct dimensions yield $\mathbf{\Phi} = [\phi(s_1 - 0) \phi(s_2 - 0), \phi(s_1 - 0) \phi(s_2 + 1), \phi(s_1 + 1) \phi(s_2 - 0), \phi(s_1 +1) \phi(s_2 + 1)]$.

\textbf{Adaptive Methods:}
This still has an exponential number of terms, however.
To remedy this, we can start with a minimal basis set and add extra resolution \textit{as and when necessary}. This can be done using two atomic operations: \textit{splitting} and \textit{combining}. The former takes a feature $\phi$ and replaces it with its children, redistributing its weight among the children, such that the value function remains unchanged. This allows each child's weight to be updated individually, adding more representational capacity. In the above example, if we were to refine $\mathbf{\Phi}_1$ in dimension $1$, we obtain the children of $\phi(s_1)$: $\phi(2s_1), \phi(2s_1 - 1), \phi(2s_1 - 2)$ and multiply each child with $\phi(s_2)$ to obtain 3 new functions, which would replace $\mathbf{\Phi}_1$.
The second operation, \textit{combining}, adds in products between different dimensions dynamically, instead of starting with a full combination. Intuitively, this allows us to start with a decoupled basis set (with $(n + 2^j)d$ terms as opposed to $(n + 2^j)^d$ for a fully coupled basis), where interactions between different state variables are ignored, and only added in when doing so would improve our approximation of the value function. For example, the initial basis set would simply be $\mathbf{\Phi} = [\phi(s_1), \phi(s_1 + 1), \phi(s_2), \phi(s_2 + 1)]$, and conjunctions could be subsequently added.

However, this raises a new question: \textit{how do we choose which functions to split and which to combine?} 
One solution is to estimate the usefulness or \textit{relevance} of a candidate function as we learn the value function.
We first define $    H(\phi, E) = \frac{T-1}{T}|| \Omega || (1 - \epsilon) \sum_{t=0}^T \epsilon^{(T - t)} E(s_t) \phi(s_t) $
where $T$ is the sample count of $\phi$ (the number of times that $\phi$ was nonzero for a given state), $|| \Omega ||$ is the size of the domain in which $\phi$ is nonzero and $\epsilon$ is a hyperparameter which controls how much weight in the moving average is put on recent samples. 
We then extend the relevance measure of \citet{geramifard2011online} to obtain the \textit{relevance} $\rho(\phi) = H(\phi, E(s_t) = \delta(s_t))$ and the \textit{observed error} $O(\phi) = H(\phi, E(s_t) = |\delta(s_t)|)$, where $\delta_t$ is the \textit{TD error} $\delta_t = R(s_{t}, a_{t}) + \gamma V(s_{t+1}) - V(s_t)$. These quantities are estimates of the inner products $\langle \phi, \delta \rangle $ and $\langle \phi, |\delta| \rangle $, respectively. Intuitively, functions with high relevances will be frequently nonzero in the presence of error and will thus reduce this error when added to the basis set.
Finally, we define $C(\phi) \dot = O(\phi) - \rho(\phi)$ for convenience.

Using the above ideas of \textit{splitting, combining} and \textit{relevance}, we first present AWR (\autoref{alg:awr}), in which we start with a full basis set at some scale and adaptively increase the scale for functions that have the highest $C$ above some tolerance. This is similar to the method proposed by \citet{li2019Adaptive} (who used Haar wavelets, although the theoretical analysis is more generally applicable), but we split based on the function relevance instead of using the coefficient magnitudes. 
The next algorithm is IBFDD (\autoref{alg:ibfdd}), which starts with a decoupled basis set and adds in interactions (i.e. products) of different basis functions as the method progresses. Finally, we combine these two to obtain the Multiscale Adaptive Wavelet Basis (MAWB), which simply starts with a decoupled basis set, and alternates between IBFDD and AWR as learning progresses---either increasing the detail in a local region or adding in useful interactions. 
\begin{minipage}[t]{0.48\textwidth}
\begin{algorithm}[H]
\footnotesize
\awr
\caption{AWR: Adaptive Wavelet Refinement}
\label{alg:awr}
\end{algorithm}
\end{minipage}
\hfill
\begin{minipage}[t]{0.48\textwidth}
\begin{algorithm}[H]
\footnotesize
\ibfdd
\caption{IBFDD: Incremental Basis Function Dependency Discovery}
\label{alg:ibfdd}
\end{algorithm}
\end{minipage}

\subsection{Preliminary Experiments}
\begin{figure}[h]
\newcommand{\fww}{0.2\textwidth}
    \centering
    \begin{subfigure}{\fww}
        \centering
        \includegraphics[width=1\linewidth]{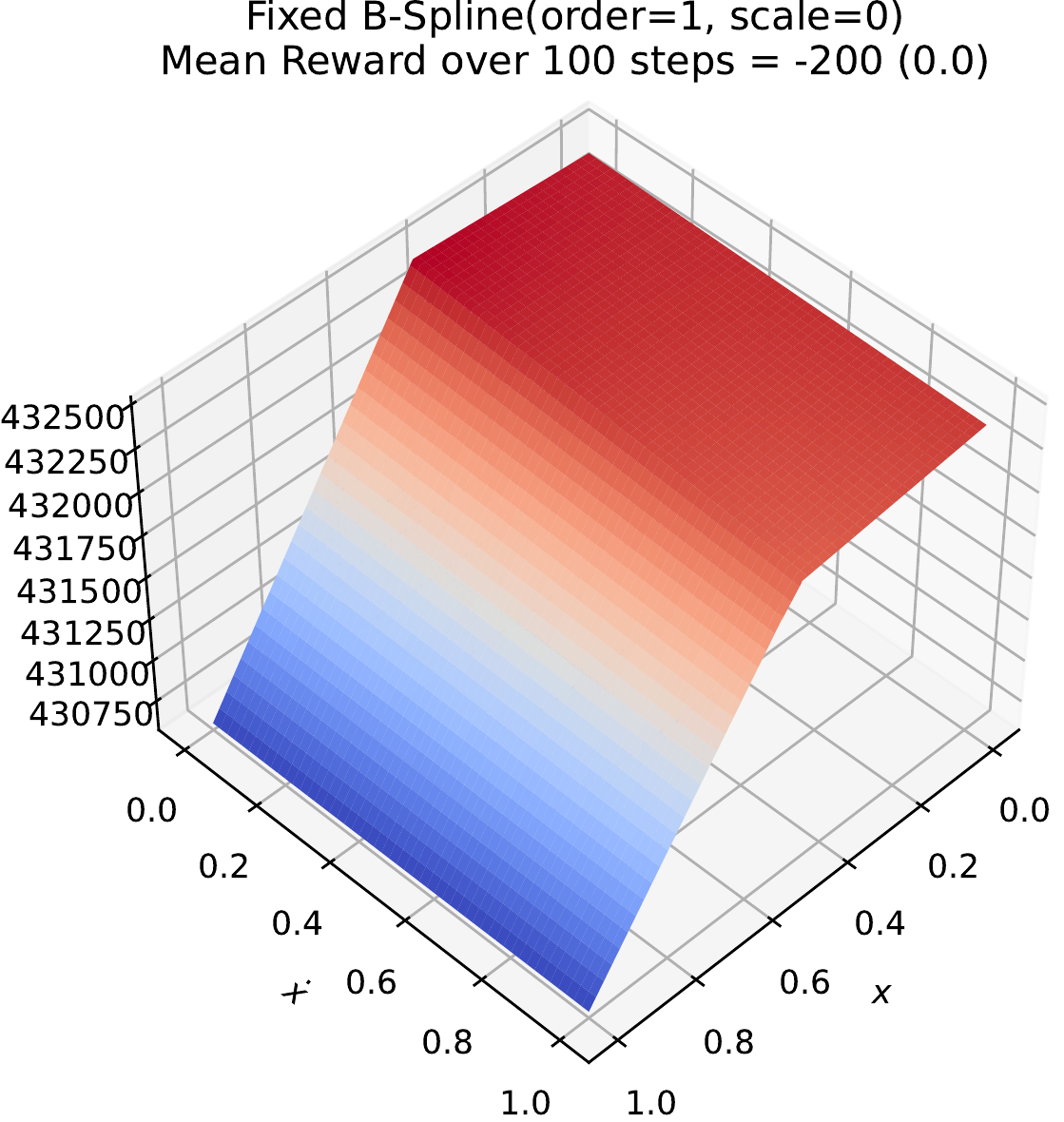}
        \caption{Fixed B-Spline}
        \label{fig:value_func_fixed_spline_10}
    \end{subfigure}
    \begin{subfigure}{\fww}
        \centering
        \includegraphics[width=1\linewidth]{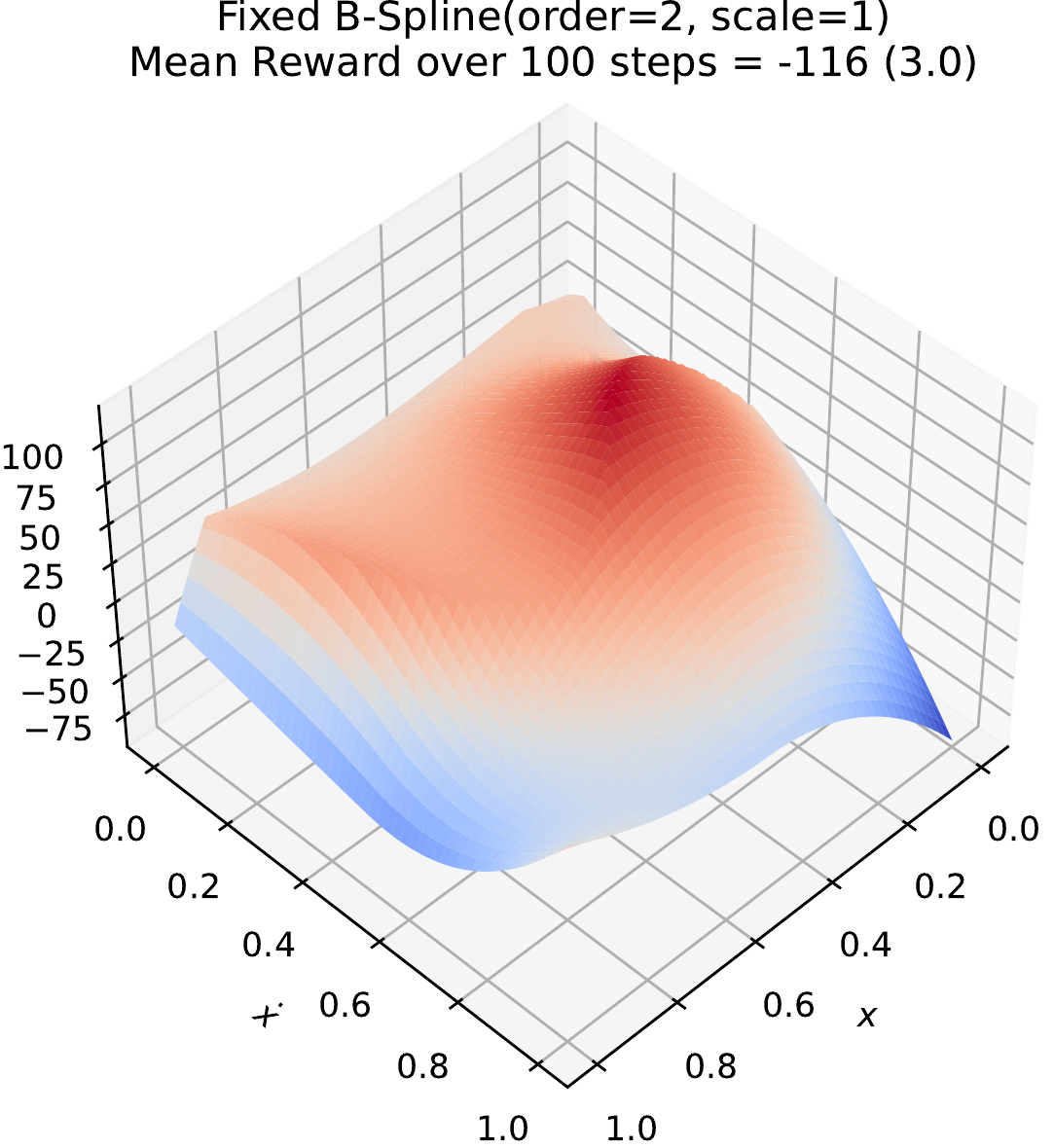}
        \caption{Fixed B-Spline}
        \label{fig:value_func_fixed_spline_21}
    \end{subfigure}
    \begin{subfigure}{\fww}
        \centering
        \includegraphics[width=1\linewidth]{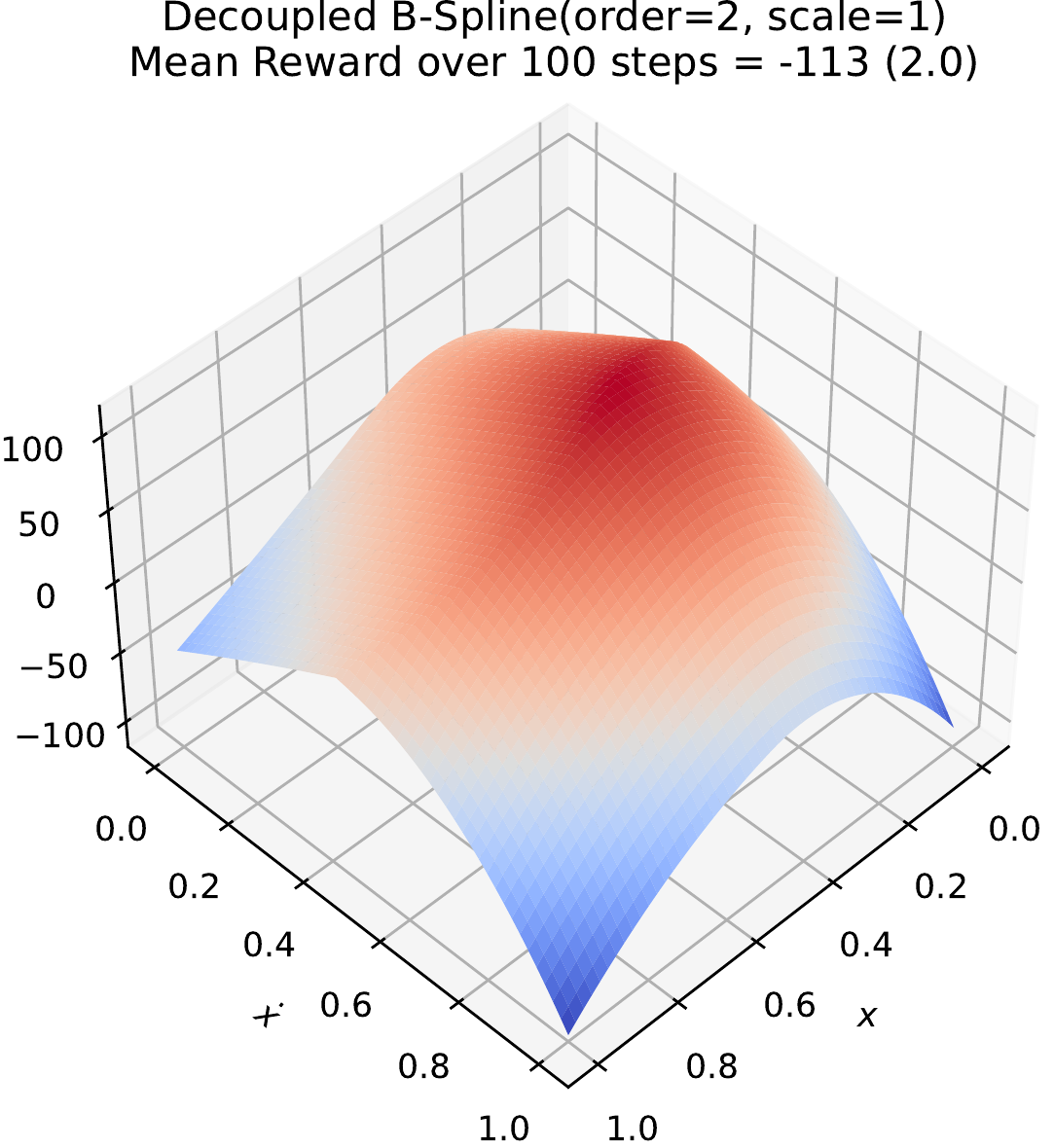}
        \caption{Fixed B-Spline, Decoupled}
        \label{fig:value_func_ibfdd}
    \end{subfigure}
    \begin{subfigure}{\fww}
        \centering
        \includegraphics[width=1\linewidth]{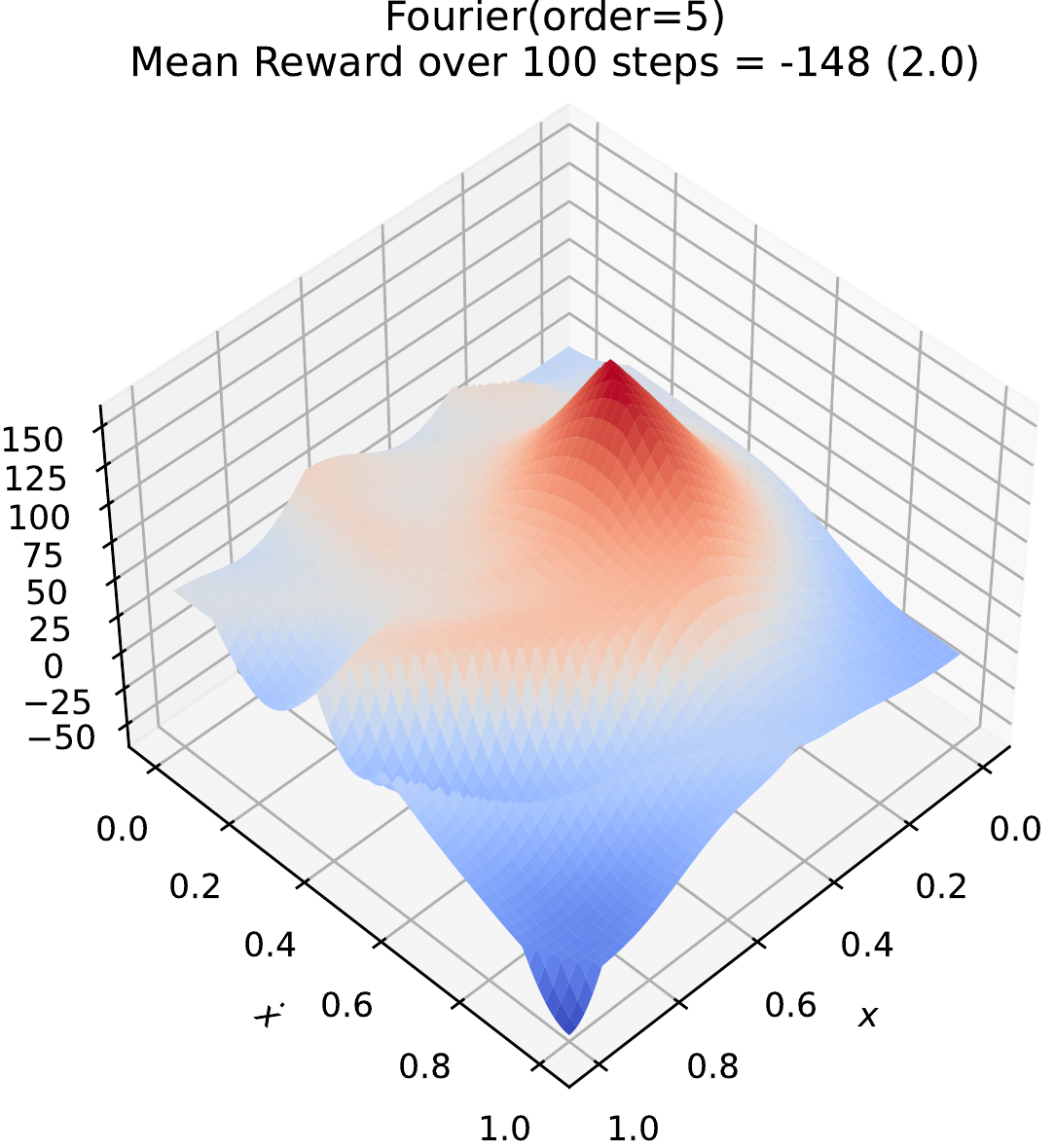}
        \caption{Fourier}
        \label{fig:value_func_fourier}
    \end{subfigure}
    \caption{Negative value functions for Mountain Car showing mean reward (standard deviation) over 10 runs of 100 episodes with the shown value function being fixed. These were achieved after performing $10,000$ episodes with a small $\alpha$. An insufficient scale as in (a) performs poorly, and the resulting value function in (b) is smoother than that of (d), while achieving improved results. The decoupled basis set also results in high performance, with a simple value surface.}
    \label{fig:all_value_funcs}
\end{figure}

In \autoref{fig:all_nonadap} we demonstrate the performance of the wavelet basis on two tasks using Sarsa($\lambda$).
We see that, for Mountain Car, the fixed B-spline basis with 36 terms in total is competitive with the Fourier basis with 36 terms.
For Acrobot, most methods do well, and again, the B-Spline basis is competitive with the Fourier basis.  The adaptive methods generally perform comparably to the fixed ones, while not needing to start with an exponentially sized basis set.  We do note, however, that using a decoupled basis (which is equivalent to MAWB with $\tau_c, \tau_s = \infty$) often outperformed a coupled one, motivating the need to investigate more complex environments.
Finally, we show example value functions in \autoref{fig:all_value_funcs}.

\newcommand{\ww}{0.39\linewidth}
\begin{figure}[h]
    \centering
    \begin{subfigure}{\ww}
    \includegraphics[width=1\linewidth]{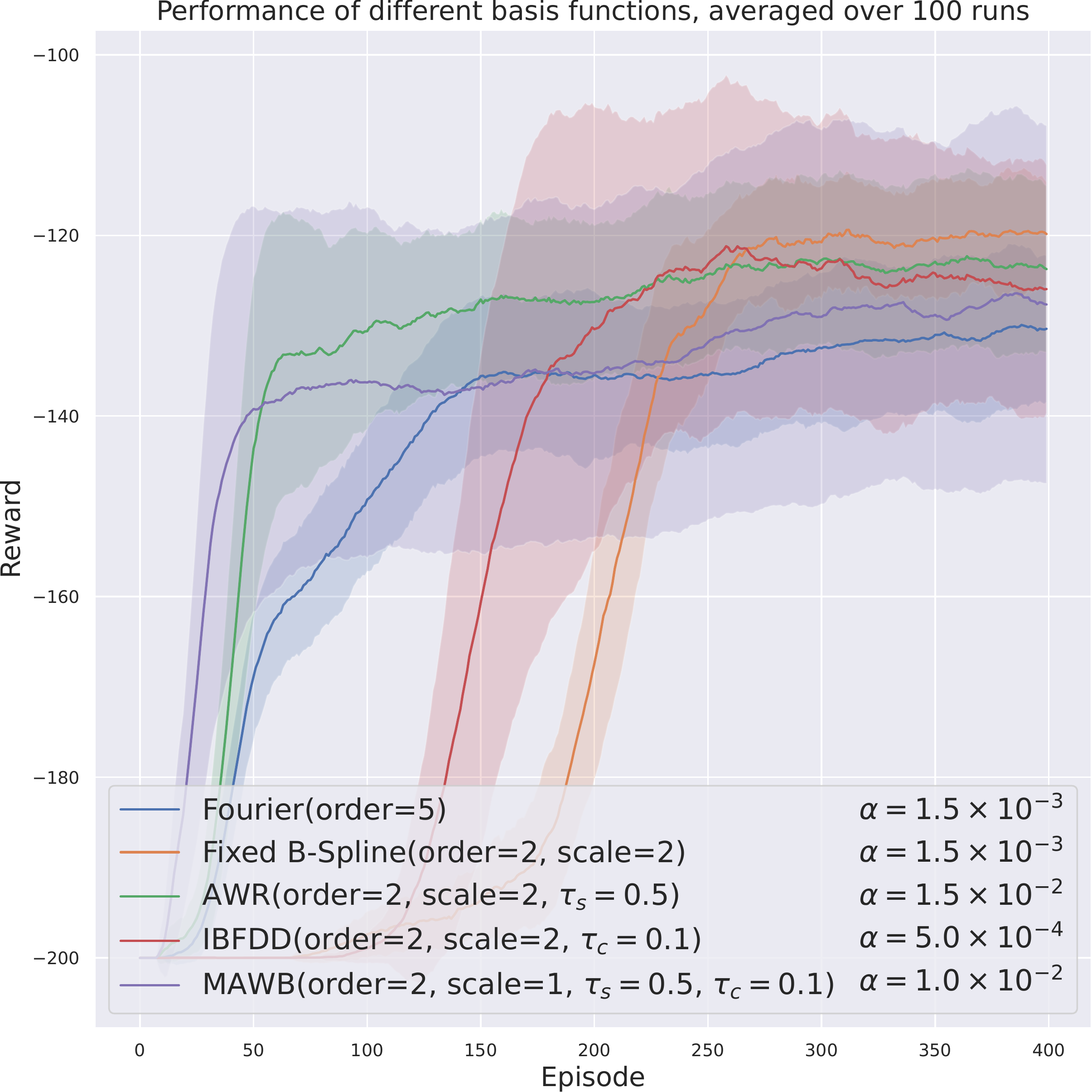}
    \caption{Mountain Car}
    \label{fig:mc_nonadaptive}
    \end{subfigure}
    \begin{subfigure}{\ww}
    \includegraphics[width=1\linewidth]{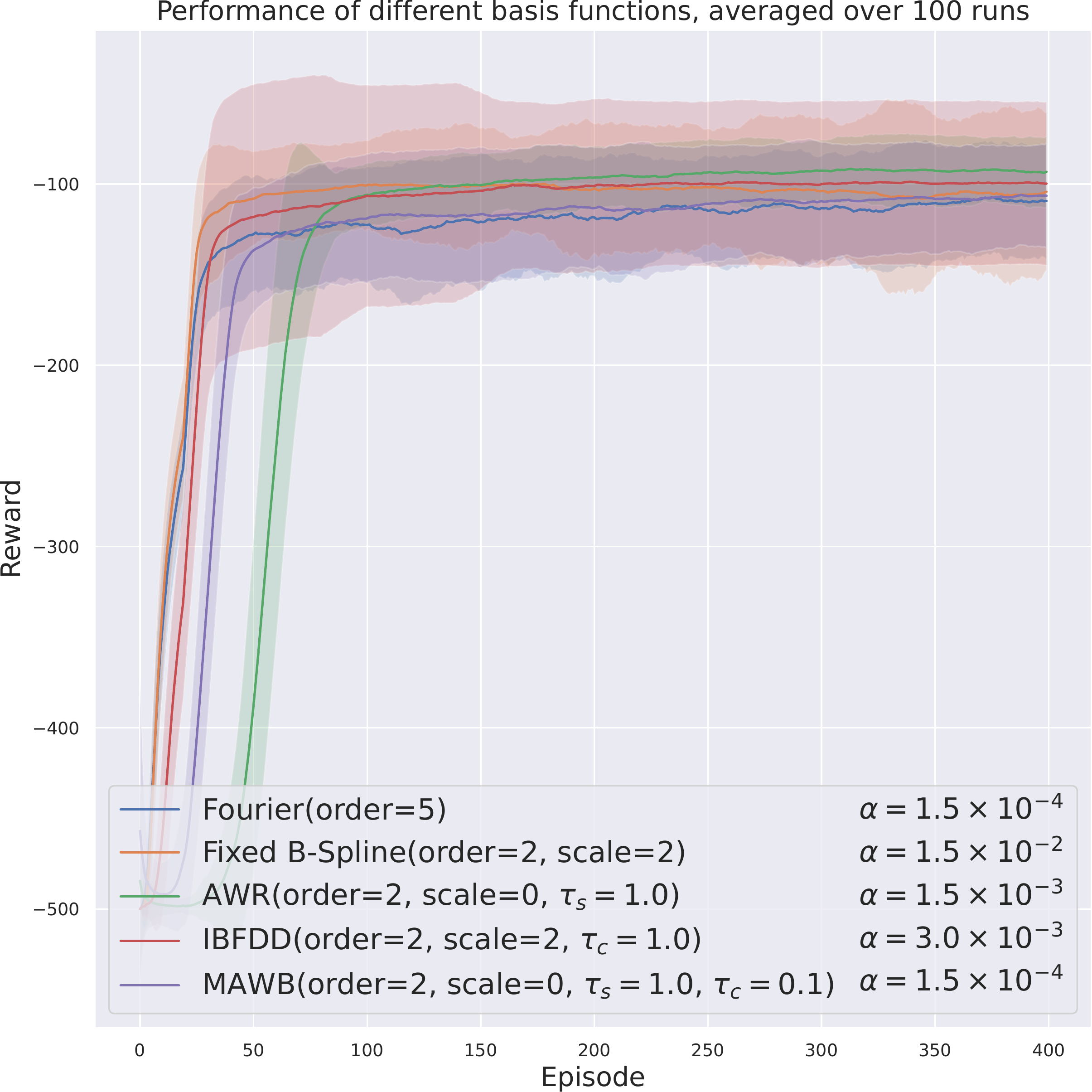}
    \caption{Acrobot}
    \label{fig:acro_nonadaptive}
    \end{subfigure}
\caption{The performance of different basis functions on (a) Mountain Car and (b) Acrobot. $\alpha$ was chosen for each basis function based on a small grid search to maximise the mean reward over the last 100 episodes.
We use Sarsa($\lambda$) ($\gamma = 1, \lambda=0.9, \epsilon=0$) and we plot the average reward over the previous 20 episodes, with standard deviation shaded. }
    \label{fig:all_nonadap}
\end{figure}

\section{Conclusion}

We introduced a linear function approximation scheme whose features are wavelet functions. 
We proved that the use of wavelets is both necessary and sufficient to obtain a refinable basis of $\mathcal{L}_2(\mathbb{R})$ that can perform multi-scale function approximation.
Preliminary experimental results demonstrate that wavelets are competitive with other fixed-basis function approximation schemes.

{\footnotesize
\bibliography{bib}
}
\end{document}